\documentclass[10pt,twocolumn,letterpaper]{article}

\usepackage[accsupp]{axessibility}  
\usepackage{iccv}
\usepackage{times}
\usepackage{epsfig}
\usepackage{graphicx}
\usepackage{amsmath}
\usepackage{amssymb}
\usepackage{algorithm}
\usepackage{algorithmic}
\usepackage{multirow}
\usepackage{xcolor}

\definecolor{citecolor}{HTML}{0071bc}
\usepackage[pagebackref=true,breaklinks=true,letterpaper=true,colorlinks,  citecolor=citecolor,bookmarks=false]{hyperref}

\makeatletter
\def\@fnsymbol#1{\ensuremath{\ifcase#1\or \dagger\or \ddagger\or
   \mathsection\or \mathparagraph\or \|\or **\or \dagger\dagger
   \or \ddagger\ddagger \else\@ctrerr\fi}}
\makeatother

\iccvfinalcopy 

\def\iccvPaperID{8702} 

\ificcvfinal\fi

\begin{document}

\title{BiViT: Extremely Compressed Binary Vision Transformers}

\author{%
  Yefei He\textsuperscript{1} 
  \quad Zhenyu Lou\textsuperscript{1}
  \quad Luoming Zhang\textsuperscript{1}
   \quad Jing Liu\textsuperscript{2}
  \quad Weijia Wu\textsuperscript{1} \\
  \quad Hong Zhou\textsuperscript{1$\dag$}
  \quad Bohan Zhuang\textsuperscript{2}\thanks{H. Zhou and B. Zhuang are corresponding authors.}
  \\[0.2cm]
  \textsuperscript{1}Zhejiang University, China \\
  \textsuperscript{2}ZIP Lab, Monash University, Australia
}
\maketitle
\ificcvfinal\thispagestyle{empty}\fi

\begin{abstract}
Model binarization can significantly compress model size, reduce energy consumption, and accelerate inference through efficient bit-wise operations. Although binarizing convolutional neural networks have been extensively studied, there is little work on exploring binarization of vision Transformers which underpin most recent breakthroughs in visual recognition. To this end, we propose to solve two fundamental challenges to push the horizon of \textbf{Bi}nary \textbf{Vi}sion \textbf{T}ransformers (BiViT). First, the traditional binary method does not take the long-tailed distribution of softmax attention into consideration, bringing large binarization errors in the attention module. To solve this, we propose Softmax-aware Binarization, which dynamically adapts to the data distribution and reduces the error caused by binarization. Second, to better preserve the information of the pretrained model and restore accuracy, we propose a Cross-layer Binarization scheme that decouples the binarization of self-attention and multi-layer perceptrons (MLPs), and Parameterized Weight Scales which introduce learnable scaling factors for weight binarization. Overall, our method performs favorably against state-of-the-arts by \textbf{19.8\%} on the TinyImageNet dataset. On ImageNet, our BiViT achieves a competitive \textbf{75.6\%} Top-1 accuracy over Swin-S model. Additionally, on COCO object detection, our method achieves an mAP of 40.8 with a Swin-T backbone over Cascade Mask R-CNN framework.
\end{abstract}
\section{Introduction}
\label{sec:intro}
Vision Transformer (ViT)~\cite{dosovitskiy2020vit} and its variants have achieved great success in a variety of computer vision tasks, such as image classification~\cite{dosovitskiy2020vit,liu2021swin,graham2021levit}, object detection~\cite{li2022exploring,fang2021yolos,carion2020detr}, semantic segmentation~\cite{zheng2021rethinking,strudel2021segmenter,chen2021transunet}, etc. However, massive parameters and calculations of the Transformer models hinder their applications on portable devices such as mobile phones. To tackle the efficiency bottlenecks, various model compression algorithms have been widely studied, such as distillation~\cite{touvron2021training,Touvron2022DeiTIR,jia2021efficient}, pruning~\cite{pan2021reduct,zhu2021vision,yu2021unified} and quantization~\cite{li2022qvit,lin2022fq,li2022ivit}. Among them, binary neural networks (BNN) ~\cite{DBLP:journals/corr/CourbariauxB16, rastegari2016xnor, bulat2019xnor+} aggressively compress weights and activations to a single bit, which delivers $32\times$ savings on memory consumption, and enables efficient $\mathrm{XNOR}$-$\mathrm{popcount}$ bit-wise operations to greatly accelerate model inference and reduce energy consumption.

\begin{figure}
    \centering
    \includegraphics[width=\columnwidth]{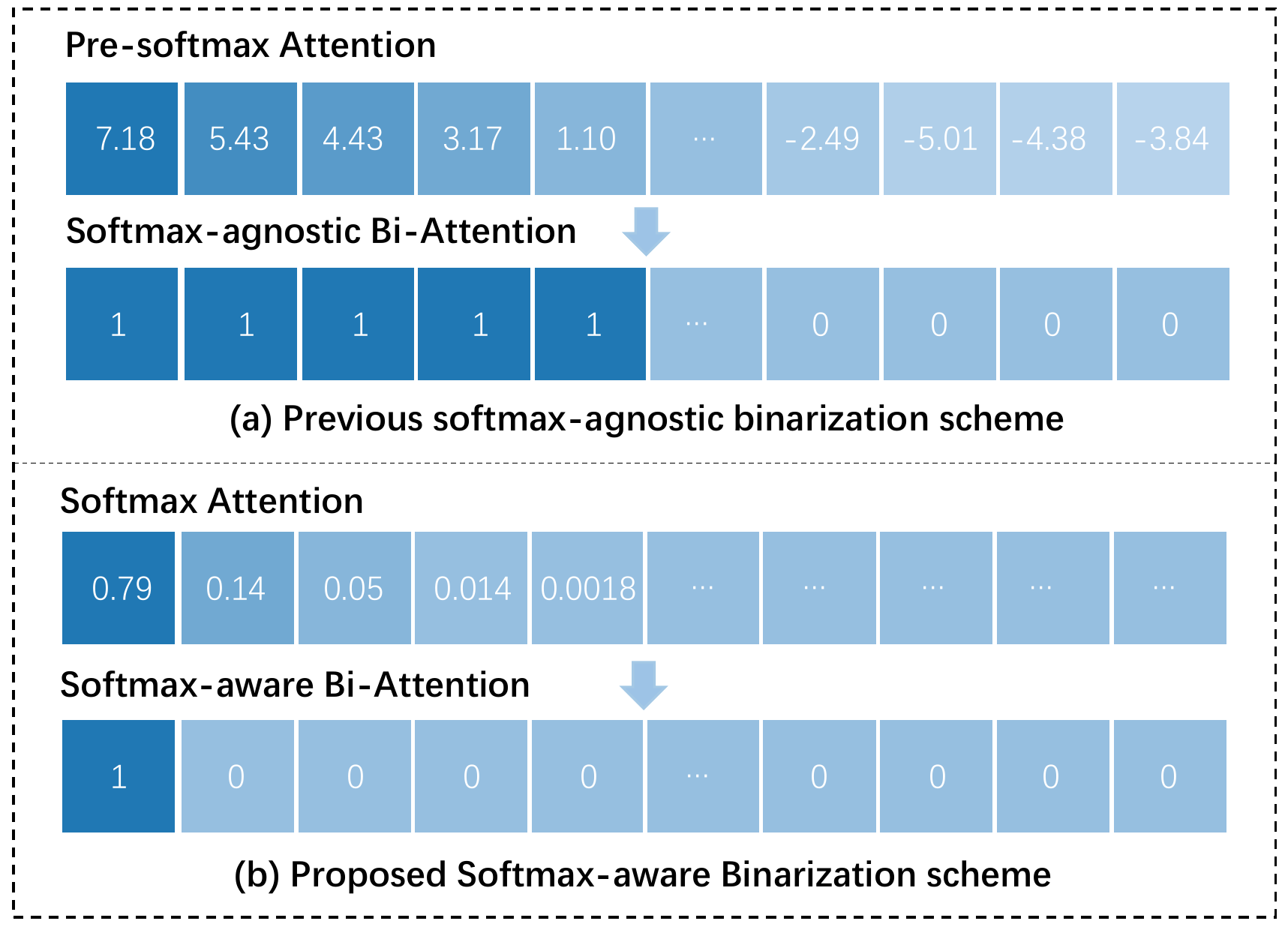}
    \caption{\textbf{An illustration of attention binarization.} Data is collected from pretrained Nest-T model and ``Bi-Attention'' denotes for binarized attentions. (a) Pre-softmax attention binarized by BiBERT~\cite{qin2022bibert}. 
    (b) Softmax attention binarized by our method. 
    }
    \label{fig:attentiondetails}
\end{figure}

However, the performance degradation restricts the wide application of BNNs, which is mainly caused by the limited representational ability and difficulty in optimization. To improve the performance of BNNs, binarized convolutional neural networks (CNNs) literature has been extensively studied to design accurate binarization functions ~\cite{rastegari2016xnor,zhou2016dorefa,bulat2019xnor+}, enhance the representation ability~\cite{liu2018bi,liu2020reactnet,zhuang2022structured} and relieve the gradient approximation error in optimization~\cite{qin2020forward,hou2016loss,bai2018proxquant}. Also, many attempts have been made in previous studies to binarize BERT~\cite{devlin2018bert} for natural language processing (NLP) tasks, such as calibrating
the attention value range mismatch~\cite{qin2022bibert}, customizing knowledge distillation and techniques in binary CNNs to Transformers~\cite{liu2022bit}. However, there are few studies on the binarization of ViTs so far. Therefore, it is highly critical and imperative to explore BiViT for diverse edge devices to infer ViT at low-latency and low-power for real-world flexibility.

In addition to the common challenges mentioned above,
binarizing Transformers presents two new exclusive
challenges. \textit{Firstly, it lacks effective methods for accurately binarizing softmax attention.}
Self-attention module aims to encode pairwise similarity between all the tokens~\cite{vaswani2017attention}, which is very different from convolutional or fully-connected layers. Specifically, the values of attention scores are all positive values between (0, 1) and exhibit long-tailed distributions after $\mathrm{Softmax}$ operation (See Figures~\ref{fig:attentiondetails} (b) and ~\ref{fig:attentiondistribution}).
In contrast, the ordinary weights have both positive and negative values and follow a bell-shaped distribution. Moreover, attention scores are generated during inference while the ordinary weights are fixed after finishing training. Consequently, the functionality and data distribution of softmax attention differ significantly from ordinary weights. As to this problem, the recent study BiBERT~\cite{qin2022bibert} proposes to maximize the information entropy of binary attention scores by applying $\mathrm{Bool}$ function on pre-softmax attentions, resulting in the balanced number of zeros and ones, as shown in Figure~\ref{fig:attentiondetails} (a). However, the softmax attention scores are actually dominated by few elements, thus the number of ones in binary attentions should be much less than the number of zeros to ensure a low quantization error (see Figure~\ref{fig:attentiondetails} (b)). 
In other words,
BiBERT follows a softmax-agnostic approach and overlooks the effect of $\mathrm{Softmax}$ on the distribution, resulting in mismatched attention score distributions before and after binarization and leading to significant quantization errors (See Table \ref{table:quanterror}). Another study BiT~\cite{liu2022bit} proposes to learn both scales and thresholds for weight and attention binarization, making them fixed during inference. While this method works well for weight binarization, it neglects the dynamics of attention scores and cannot adapt well to the changing distribution of attentions, as this approach remains softmax-agnostic at inference time as well.

\begin{figure}
    \centering
    \includegraphics[width=\columnwidth]{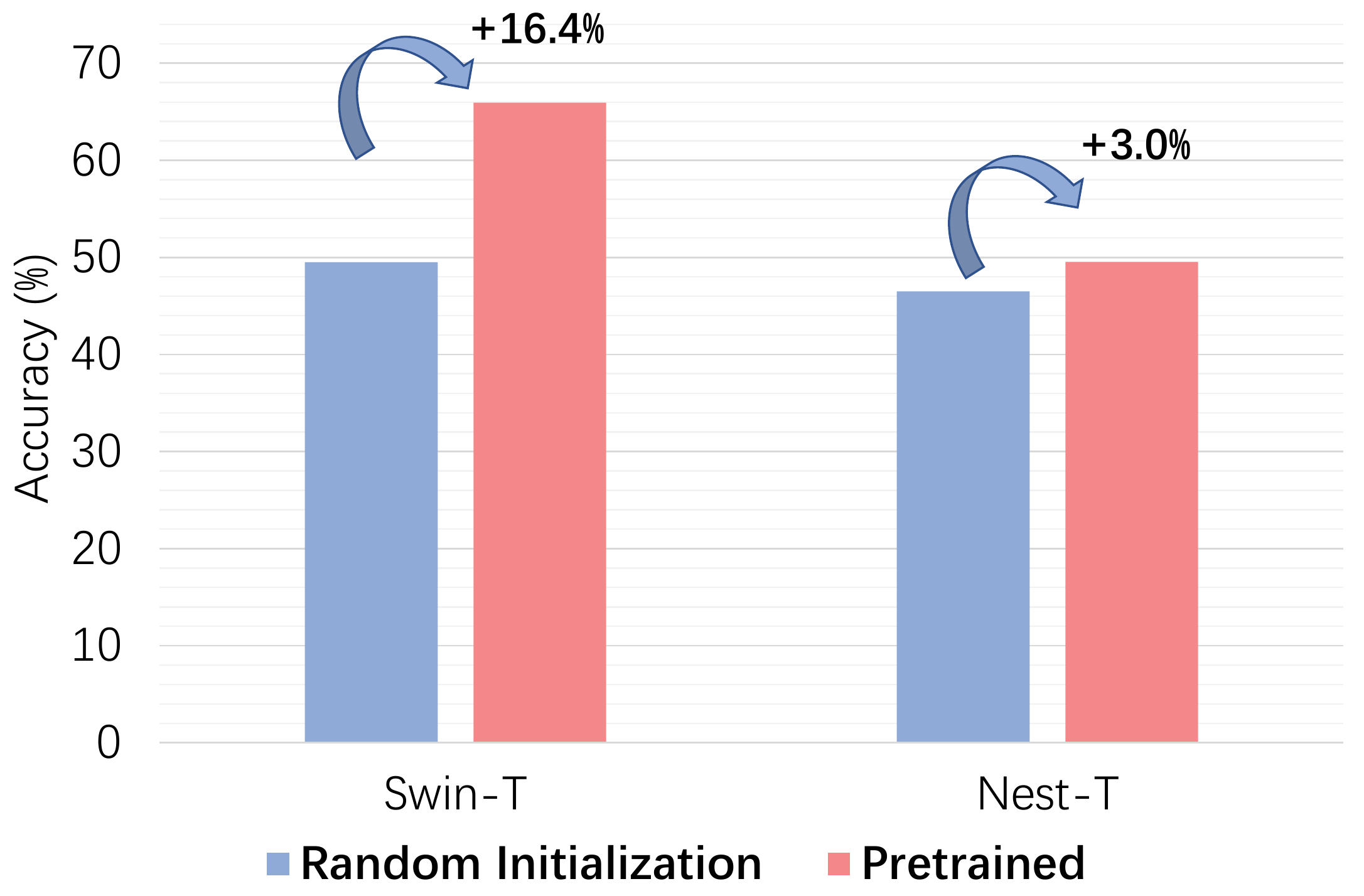}
    \caption{\textbf{Impact of pretrained model when binarizing Transformers.} The experiment is conducted on TinyImageNet dataset. Initiating Transformers from the pretrained models greatly boosts the accuracy.}
    \label{fig:pretrainimpact}
\end{figure}

\textit{Secondly, how to preserve the information in the pretrained ViTs during binarization is under explored.} Unlike binary CNNs that perform well when training from scratch~\cite{qin2020forward,tu2022adabin,liu2020reactnet}, we observe that BiViTs heavily rely on pretrained models and are sensitive to quantization, as shown in Figure~\ref{fig:pretrainimpact}.
Even if the initial weights are derived from the pretrained model, directly binarizing all parameters still causes a huge loss of pretrained information, which then leads to a severe performance drop. Also, the loss of pretrained information is difficult for Transformers to recover through quantization-aware training (QAT). In particular, MLP modules account for nearly half of the computations and parameters within a Transformer~\cite{liu2022ecoformer}. They are mainly composed of
$1 \times 1$ convolutions, which are widely recognized to be difficult to 
binarize
due to the limited representational capability~\cite{zhuang2022structured,garg2021confounding,bulat2020high}. 
Therefore, the effective binarization of softmax attention and the retention of information from pretrained models remain open questions. 


To reduce the quantization error in binarizing attentions, we first analyze the long-tailed distribution of softmax attention scores and discover their differing patterns
across
different attention vectors. To adaptively search the optimal thresholds for binarization, we propose an optimization algorithm based on sparse coding and coordinate descent, and further propose an efficient approximation called Softmax-aware Binarization (SAB) to avoid conducting the optimization on each forward pass. Moreover, to retain pretrained information and further enhance the model representational capability, we then propose Cross-layer Binarization (CLB) to decouple the quantization of self-attention and MLPs to avoid mutual interference and introduce Parameterized Weight Scales (PWS) for weights binarization. 
To our best knowledge, we are the pioneering work to 
probe
binarizing Transformers for vision tasks. 

In summary, our contributions are as follows:
\vspace{-0.5em}
\begin{itemize}
 \setlength{\itemsep}{1pt}

\item We are the pioneering work that explores binary vision Transformers, a demanding recipe for efficient ViT inference.
\item We design a Softmax-aware Binarization scheme for the self-attention module, which adapts to the long-tailed attention scores distribution and greatly reduces the quantization error.
\item We propose Cross-layer Binarization and Parameterized Weight Scales to retain pretrained information and further enhance the representational ability of BiViTs, improving 
convergence and accuracy.
\item 
Experiments on TinyImageNet and ImageNet for image classification, and COCO for object detection, demonstrate that it consistently outperforms current state-of-the-arts by large
margins, serving as a strong baseline for future research. 
\end{itemize}
\section{Related Work}
\subsection{Vision Transformers}
Transformer~\cite{vaswani2017attention} is initially proposed to process long sequences in NLP tasks. ViT~\cite{dosovitskiy2020vit} first adapts Transformers to vision tasks by splitting images into patches and processing them as token sequences. 
DeiT~\cite{pmlr-deit-touvron21a} further improves the data efficiency of vision Transformers. Benefiting from the global receptive field and the powerful long-range modeling capabilities of 
self-attention,
ViT demonstrates promising performance against CNN counterparts. Many follow-up works are proposed to explore hierarchical structures~\cite{liu2021swin,wang2021pyramid,zhang2022nested}, insert the convolutional inductive bias~\cite{li2021localvit,guo2022cmt,dai2021coatnet} and apply ViTs to various vision tasks~\cite{zhang2022topformer,zeng2021improving,fang2021yolos}.
However, the inference speed of ViTs is generally slower than that of CNNs in practical applications~\cite{li2022efficientformer}. The reasons mainly include the lack of special optimization (such as Winograd~\cite{liu2018winograd} for convolutional layers) and the quadratic computational complexity of the self-attention module.
To reduce the computational complexity of ViTs, many methods have been proposed, including linear complexity attention~\cite{shen2021linear,cai2022efficientvit,wang2022pvt}, network pruning~\cite{Hou_2022_CVPR,yang2021nvit,CHASING_SPARSITY_2021} 
and quantization~\cite{li2022qvit, lin2022fq,liu2021post}. However, current Transformer quantization literature mainly focuses on fixed-point quantization, either through Quantization-Aware Training (QAT)~\cite{li2022q,li2022qvit} or Post-Training Quantization (PTQ)~\cite{liu2021post,lin2022fq,yuan2021ptq4vit}. Research on ternary or binary quantization remains to be studied.

\subsection{Binary Neural Networks}
BNNs seek to quantize both weights and activations to 1-bit, which greatly reduces the complexity of the model. The binarization of models usually requires QAT to restore accuracy. To overcome the non-differentiability of quantizer during training and the limited representational capacity, many methods have been proposed to help binarize CNNs, such as binary-friendly model structures~\cite{liu2018bi,liu2019circulant,zhu2019benn,mishra2017wrpn, bulat2020high,bethge2021meliusnet}, knowledge distillation~\cite{liu2020reactnet,mishra2017apprentice,martinez2020training}, gradient approximation~\cite{he2022binarizing,qin2020forward,zhang2022root,ding2022ie}, 
optimizer selection~\cite{liu2021adam,alizadeh2019systematic,courbariaux2016binarized}, $\etc$  Although some of them are also effective for Transformer models, as analyzed in BiT~\cite{liu2022bit}, methods 
targeting on
binary Transformers still need to be developed to relieve accuracy degradation.

The literature closely related to our work includes BinaryBERT~\cite{bai2021binarybert}, BiBERT~\cite{qin2022bibert} and BiT~\cite{liu2022bit}. They 
propose techniques to improve 
the binarization of BERT~\cite{devlin2018bert} variants, 
such as
customized binarization functions and model distillation, and evaluate them on NLP tasks. However, none of these methods are evaluated on vision tasks. In the following sections, we will migrate these methods to DeiT~\cite{pmlr-deit-touvron21a}, Swin~\cite{liu2021swin} and NesT~\cite{zhang2022nested} as our baselines to test their performance and analyze
the key challenges.
Then we propose to improve BiViT's performance with our 
SAB, CLB and PWS. 
To the best of our knowledge, we are the pioneering work 
studying
BiViTs.

\section{Method}
\subsection{Preliminaries} \label{sec:preliminaries}
Generally, standard BNNs follow ~\cite{rastegari2016xnor} to use $\mathrm{Sign}$ function to binarize weights and activations to \{-1, +1\}, and exploit Straight-Through Estimator (STE)~\cite{bengio2013estimating} to overcome the non-differentiability of the $\mathrm{Sign}$ function, as follows:
\begin{gather}
{\hat{x} = \mathrm{Sign}}(x)=\left\{
\begin{aligned}
+1&, \mathrm{if}\  x\ge0 \\
-1&, \mathrm{otherwise},
\end{aligned}
\right.
\end{gather}
\vspace{-1em}
\begin{gather}
    \frac{\partial \mathcal{L}}{\partial x}\approx\left\{
    \begin{aligned}
    \frac{\partial \mathcal{L}}{\partial \hat{x}}&, \mathrm{if}\ \lvert x \rvert \le1 \\
    \noindent 0&, \mathrm{otherwise}.
    \end{aligned}
    \right.
\end{gather}

To approximate the full-precision $\mathbf{x}\in\mathbb R^{n}$, BNNs further introduce a scaling factor $\alpha \in\mathbb R^{+}$ to reduce the quantization error:

\begin{equation}
\label{eq:scale}
\alpha = \frac{{\lVert \mathbf{x} \rVert}_{\ell_1}}{n},\quad
\mathbf{x} \approx \alpha \hat{\mathbf{x}}.
\end{equation}

However, 
binarization using $\mathrm{Sign}$ can be problematic
in Transformers and be very different from CNNs.
Specifically, 
in the self-attention mechanism~\cite{vaswani2017attention}, 
attention is calculated as
\begin{gather}
\text{Attention}(\mathbf{Q}, \mathbf{K}, \mathbf{V}) = \mathrm{Softmax}\left(\frac{\mathbf{Q}\mathbf{K}^T}{\sqrt{d_k}}\right) \mathbf{V},
\end{gather}
where $\mathbf{Q}$, $\mathbf{K}$, $\mathbf{V}$ are query, key and value matrics respectively, and $d_k$ is the dimension of the key. In the following, we will denote softmax attention vector (\ie, one row of the softmax attention matrix)  as $\mathbf{a}_s$ and pre-softmax attention vector as $\mathbf{a}_p$.

According to the definition of $\mathrm{Softmax}$, the results are non-negative and they will all be $+1$ after $\mathrm{Sign}$ function. In order to solve the problem of value range mismatch, BiBERT~\cite{qin2022bibert} proposes to use the $\mathrm{Bool}$ function to binarize pre-softmax attention scores to \{0, 1\}, which is defined as:
\begin{gather}
\mathrm{Bool}(a_p)=\left\{
\begin{aligned}
1&, \mathrm{if}\ a_p\ge 0 \\
0&, \mathrm{otherwise}.
\end{aligned}
\right.
\end{gather}
However, this will lead to huge quantization errors since $\mathrm{Softmax}$ is totally discarded, as we will analyze in Section~\ref{sec:sab}.

\subsection{Softmax-aware Binarization} 
\label{sec:sab}
\begin{figure}
    \centering
    \includegraphics[width=0.8\columnwidth]{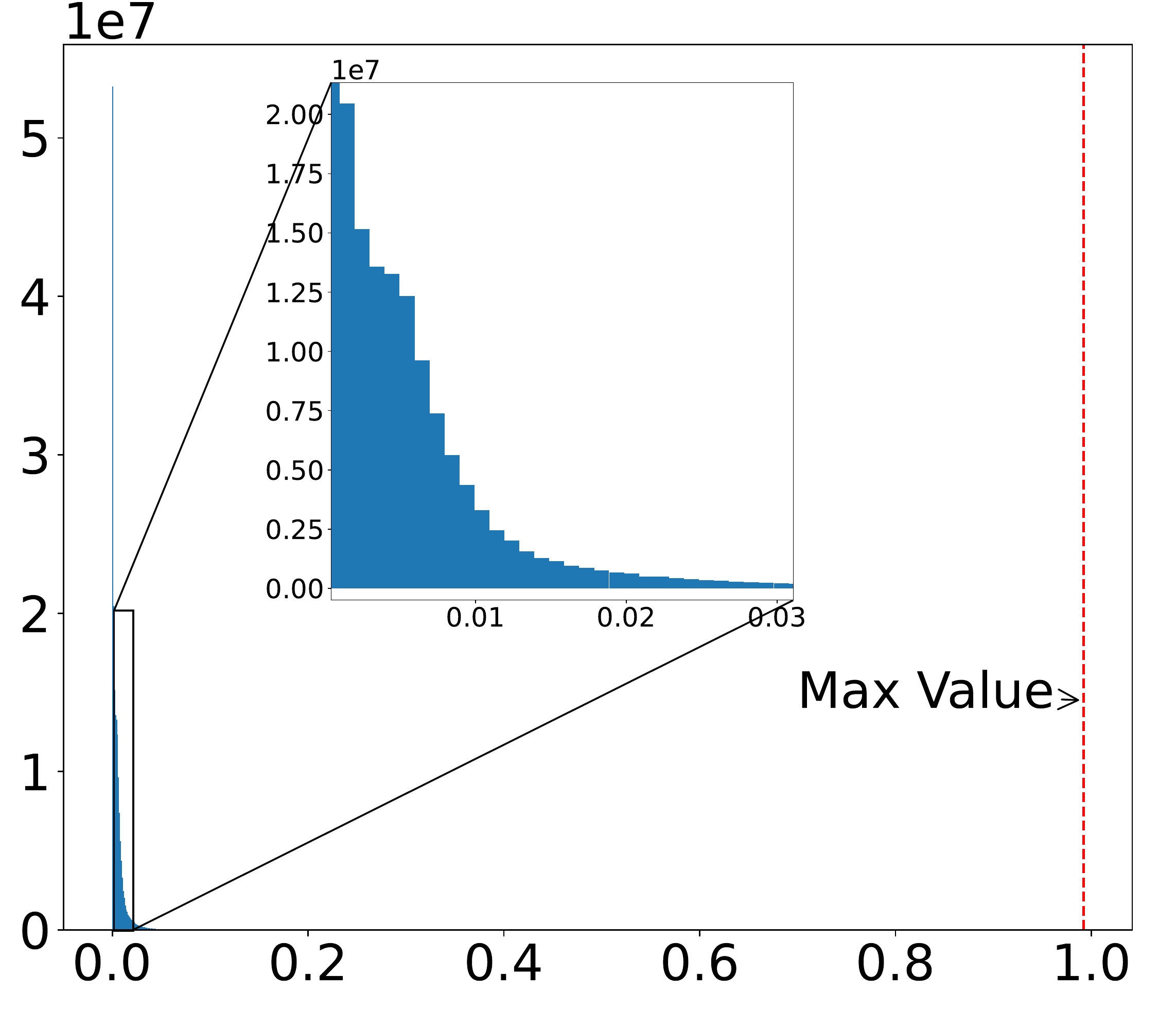}
    \caption{\textbf{The long-tailed distribution of attention scores.} Most attention scores are around zero but the maximum values can reach $0.99$.}
    \label{fig:attentiondistribution}
\end{figure}

Self-attention 
is designed to model global relationships among different patches (tokens) and focuses on important token pairs. 
Figure~\ref{fig:attentiondistribution} presents the distribution of attention scores in the pretrained NesT-T~\cite{zhang2022nested} model. 
We observe that
after $\mathrm{Softmax}$ operation, attention scores follow a long-tailed distribution and more than $99.5\%$ of them are less than $0.05$, which is highly sparse. 
To further investigate this distribution, we take a deep look at an actual attention vector 
from the NesT-T pretrained model. As shown in Figure~\ref{fig:attentiondetails} (a), if we follow BiBERT~\cite{qin2022bibert} and use $\mathrm{Bool}$ function to binarize pre-softmax attentions, nearly half of the attention scores are set to $1$, indicating they have the same contribution to binarization, which is inconsistent with the actual distribution of softmax attention scores where only few values dominate (see Figure~\ref{fig:attentiondetails} (b)).
BiT~\cite{liu2022bit} proposes learnable thresholds for binarization but it is fixed after training, while distributions of attention scores can vary with different input images and may differ between each attention vector, leading to a suboptimal solution. 

In order to reduce the quantization error while binarizing softmax attention scores,
we argue that the ideal binarization method 
should
satisfy the following two properties: 1) The proportion of activated scores (set to $1$) in binary attentions should be smaller compared to directly using the $\mathrm{Bool}$ function. 
As shown in Figure~\ref{fig:attentiondistribution}, most values are around 0, which barely contributes to the result during calculation, and only a few significant values are considered.
2) The activation thresholds should not be a fixed value. 
$\mathrm{Softmax}$ is operated on every row-wise attention vector and different softmax attention vectors follow distinct distributions. For example, the maximum value of some of them can reach $0.99$, while the others are only about $0.05$. Empirically, even though most softmax attentions are dominated by only a few elements, the thresholds to activate should be different across all attention vectors.

To achieve this, the key is to find the optimal threshold $T^*$ for binarizing each softmax attention vector (\ie, $T^*$ is different for each row). Inspired by sparse coding~\cite{lee2006sparse} and LQ-Nets~\cite{zhang2018lq}, we formulate the quantized attention vector $\mathbf {a}_q \in\mathbb R^{n}$  by the inner product between a basis vector $\mathbf v \in\mathbb R^{k}$ and the binary encoding vector $\mathbf b \in \{0,1\}^{k \times n}$:
\begin{equation} \label{eq:formulatex}
    \begin{aligned}
\mathbf {a}_q = \mathbf v^{T} \mathbf b,
    \end{aligned}
\end{equation}
where $k$ is the target bitwidth. Then the optimization problem can be formulated as:

\begin{equation} \label{eq:optimization}
  \begin{aligned}
\mathbf{v^*},\mathbf{b^*}=\mathop{\arg\!\min}\limits_{\mathbf{v}, \mathbf{b}} {\lVert{\mathbf v^{T} \mathbf b -\mathbf{a}_s}\rVert}_{2}^2 ,\quad s.t.\  \mathbf{b}\in\mathbb \{0,1\}^{k \times n}.
\end{aligned}
\end{equation}
In this paper, the bitwidth $k$ is set to 1, thus the basis vector $\mathbf v$ becomes a scalar $v$. 
However, with both $v$ and $\mathbf b$ to be solved, brute-force search can be computationally expensive. Instead, the optimization problem can be efficiently solved in a coordinate descent approach.
Specifically, we alternatively optimize the basis $v$ and binary encoding vector $\mathbf b$ while keeping another fixed:
\\ \hspace*{\fill} \\
\noindent\textbf{Update $v$:} With the fixed binary encoding vector $\mathbf b$, the optimization problem will degenerate to a special case of linear regression. Therefore, the optimal $v$ can be derived by:
\begin{equation} \label{optimalv}
v^* = \frac{\mathbf{a}_s \cdot \mathbf{b}}{{\lVert \mathbf{b} \rVert}_{2}^2} ,
\end{equation}
where $\cdot$ represents the dot product between two vectors.

\noindent\textbf{Update $\mathbf b$:} Since we get the optimal $v$ with Eq.~(\ref{optimalv}), the two values for binarization becomes $\{0, v^*\}$. Then the optimal threshold can be simply calculated as:
\begin{equation} \label{eq:calculateT}
T^* = \frac{0+v^*}{2} .
\end{equation}
Then we binarize the full-precision softmax attention vector with the threshold $T^*$ to update the binary encoding $\mathbf b$:
\begin{equation} \label{eq:binarizeattention}
\mathbf{b^*} = \mathrm{Bool}(\mathbf{a}_s-T^*).
\end{equation}

The coordinate descent optimization process is summarized in Algorithm~\ref{algor:optimization}. After $N$ iterations, the quantization error between binary attentions and full-precision softmax attentions decreases significantly, as shown in the second row of Table~\ref{table:quanterror}.

\begin{algorithm}[hbpt]
	\caption{The coordinate descent optimization.} 
	\label{algor:optimization}
	\small
	\begin{algorithmic}[1]
		\STATE \textbf{Input}: softmax attention vector $\mathbf a_s$
        \STATE \textbf{Output}: the basis scalar $v$, the binary encoding $\mathbf{b}$
		\STATE {\textbf{Procedure:}}
		\STATE \quad Initialize $\mathbf{b}^{(0)}=\mathrm{Sign}(\mathbf a_s)$\\
		\STATE \quad \textbf{for} $t$ = $1$ $\rightarrow$ $N$
  \textbf{do}\\
        \STATE \quad \quad Update $v^{(t)}$ with $\mathbf a_s$ and $\mathbf{b}^{(t-1)}$ per Eq. (8)
        \STATE \quad \quad Update $\mathbf{b}^{(t)}$ with $\mathbf a_s$ and $v^{(t)}$ per Eqs. (9) and (10)
        \STATE \quad \textbf{end for}\\
	\end{algorithmic}
\end{algorithm}

\begin{table}[h] 
\caption{Quantization error under different methods. We set $N=5$ in practice.}
\label{table:quanterror}
\centering
\scalebox{0.8}
{
\begin{tabular}{cc}
\hline
Method                   & Quantization Error \\ \hline
BiBERT                   & 0.683              \\
Optimal $T^*$                & 2.58e-05           \\
Approximate $T$            & 2.72e-05           \\
Approximate $T$ w/o scales & 0.141              \\ \hline
\end{tabular}}
\end{table}

Although this optimization strategy minimizes the quantization error, it is 
not practical
to optimize each generated attention vector during inference. Besides, we calculate an optimal $v^*$ for each attention vector, which introduces an extra computational burden. 

\begin{figure}[t]
    \centering
    \includegraphics[width=0.8\columnwidth]{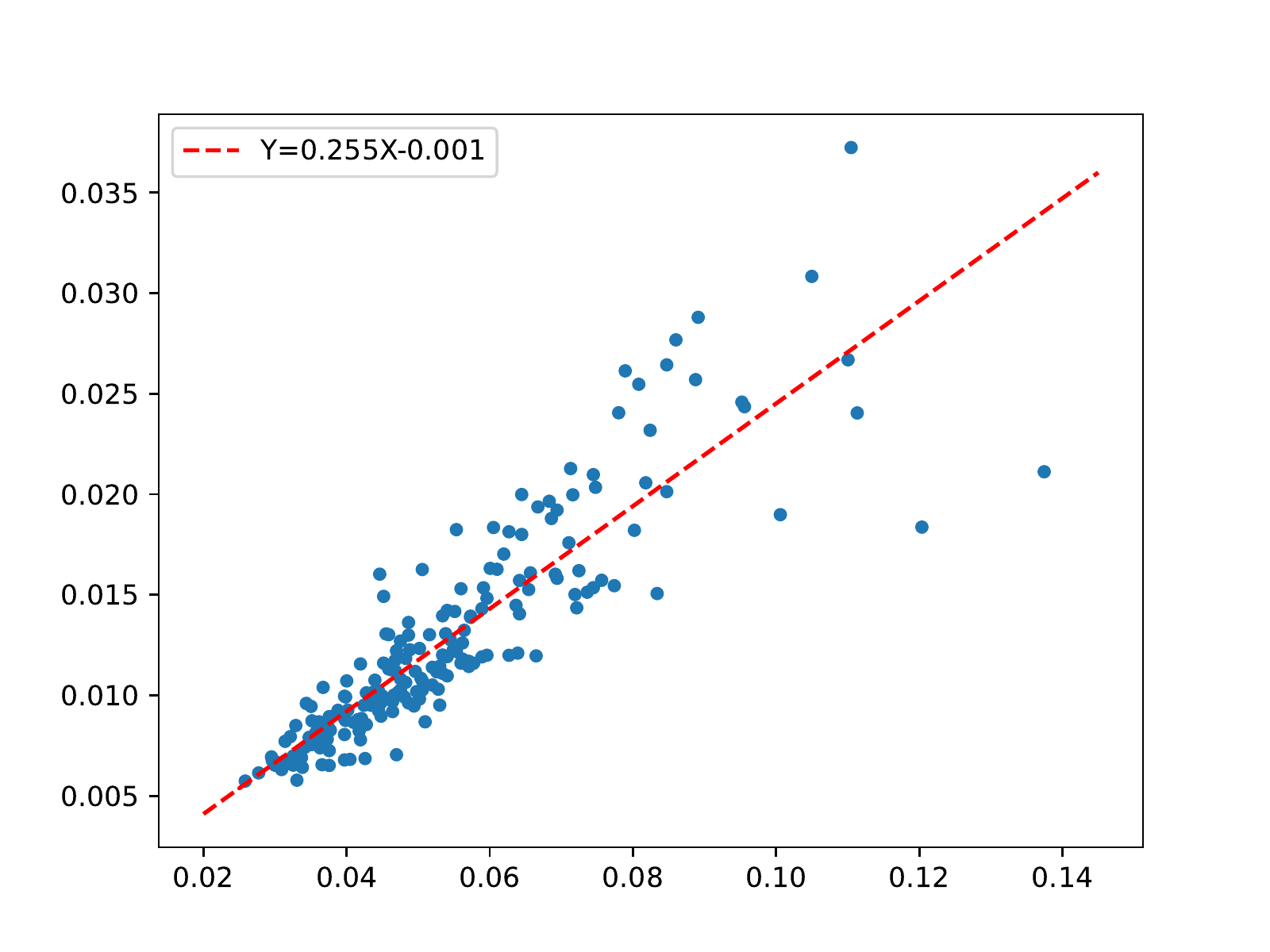}
    \caption{\textbf{Relation between the maximum values of softmax attention vectors (X-axis) and optimal thresholds $T^*$ (Y-axis).} Blue dots are the maximum values of softmax attention vectors sampled from the pretrained NesT-T model and red dashed line represents the result of linear regression on these attention scores. For simplicity, we remove the bias term in the following analysis.}
    \label{fig:relationva}
\end{figure}

To simplify the optimization process and reduce computational complexity,
we seek to establish a relationship between the optimal thresholds $T^*$ (calculated by Eq.~(\ref{eq:calculateT})) and the distribution of the softmax attention scores. In practice, we sample $128$ images from ImageNet~\cite{deng2009imagenet} dataset, obtain 
$10^4$
attention vectors and calculate the corresponding optimal thresholds $T^*$ with the proposed method. As illustrated in Figure~\ref{fig:relationva}, we observe that the optimal thresholds $T^*$ show a strong correlation with the maximum values of the softmax attention vectors.
Therefore, we estimate a coefficient $\beta$ via linear regression with these sampled data to approximate thresholds $T$ for each attention vector and thus accelerate the optimization:
\begin{equation} \label{eq:alpha}
T = \beta  \mathrm{Max}(\mathbf{a}_s). 
\end{equation}
Experimental result demonstrates that this approximation barely increases the quantization error, as shown in the third row of Table~\ref{table:quanterror}. 

However, compared with the previous methods~\cite{qin2022bibert,bai2021binarybert,liu2022bit}, multiplying the basis scalar $v$ by the binary encoding vector $\mathbf{b}$ to get $\mathbf{a}_q$ (as defined in Eq.~(\ref{eq:formulatex})) for each attention vector still introduces extra computation. To keep the same computational complexity as previous methods, we make a second approximation and further discard the 
basis scalar
$v$.
In this case, the quantization error is shown in the fourth row of Table~\ref{table:quanterror}. It should be noted that this step is simply a trade-off between accuracy and complexity. 
By default, we use the algorithm with two approximations for experiments.

Note that there are also many intuitive schemes for approximating the $\mathrm{Softmax}$ operation, such as Top-$N$ algorithm with a learnable parameter $N$ and learnable thresholds $T$ proposed in BiT~\cite{liu2022bit}.
For Top-$N$ algorithm, it can take heavy computations to obtain $N$ and needs to find $N$ maximum values during inference. Instead, we only need to find the maximum value through the proposed twice approximation algorithm, which is faster and barely increases error. Compared with BiT~\cite{liu2022bit}, our method calculates the thresholds dynamically and can adapt well to the attention distributions during inference.

In the backward pass, BiBERT employs the STE to propagate gradients from binary attentions $\mathbf{a}_q$ to pre-softmax attentions $\mathbf{a}_p$ \emph{without} considering the $\mathrm{Softmax}$ operation:
\begin{equation} \label{eq:bibertSTE}
\setlength{\abovedisplayskip}{3pt}
\setlength{\belowdisplayskip}{3pt}
\frac{\partial \mathcal L}{\partial\mathbf a_p} \approx\frac{\partial \mathcal L}{\partial \mathbf{a}_q} .
\end{equation} 
In contrast, our proposed method employs the STE to propagate gradients from binary attentions $\mathbf{a}_q$ to softmax attentions $\mathbf{a}_s$, making it \emph{Softmax-aware} during back-propagation: 
\begin{equation} \label{eq:softmaxSTE}
\setlength{\abovedisplayskip}{3pt}
\setlength{\belowdisplayskip}{3pt}
\frac{\partial \mathcal L}{\partial\mathbf a_p} = \frac{\partial \mathcal L}{\partial\mathbf a_s} \frac{\partial\mathbf a_s} {\partial\mathbf a_p} \approx\frac{\partial \mathcal L}{\partial \mathbf{a}_q} \frac{\partial\mathbf a_s} {\partial\mathbf a_p}.
\end{equation} 
We will further show in Section~\ref{abla:sab} that this greatly helps the training of BiViTs.

Overall, the training process of the proposed SAB is summarized in Algorithm~\ref{algor:sab}.

\begin{algorithm}[htbp]
	\caption{SAB for self-attention modules} 
	\label{algor:sab}
	\small
	\begin{algorithmic}[1]
		\STATE \textbf{Input}: softmax attention scores $\mathbf a_s \in\mathbb R^{n}$
        \STATE \textbf{Output}: binary attention scores $\mathbf{a}_q \in\mathbb R^{n}$
		\STATE {\textbf{Forward propagation:}}
		\STATE \quad Approximate the thresholds 
  $T$ by the maximum value of attentions with Eq.~(\ref{eq:alpha}).\\
		\STATE \quad Binarize attentions with thresholds $T$ by Eq.~(\ref{eq:binarizeattention}).\\
		\STATE {\textbf{Backward propagation:}}
		\STATE \quad Calculate the gradients \wrt $\mathbf a_s$ with Eq.~(\ref{eq:softmaxSTE}).\\
	\end{algorithmic}
\end{algorithm}

\subsection{Information Preservation}
\subsubsection{Cross-layer Binarization} \label{sec:cr}
Compared with binary BERT~\cite{qin2022bibert,liu2022bit,bai2021binarybert}, we find that BiViTs are 
more
difficult to optimize. 
To 
justify
this, we directly migrate BiBERT~\cite{qin2022bibert} to Swin-T~\cite{liu2021swin} and NesT-T~\cite{zhang2022nested} to 
evaluate
its performance on image classification tasks.
The results are shown in Table~\ref{tinyimgnetresult}. We observe that its accuracy degradation in image classification tasks can reach $40\%$ on TinyImageNet dataset, 
which indicates that vanilla BiViTs cannot make good use of the pretrained information and is difficult to optimize on vision tasks.

\begin{figure}[htbp]
\begin{minipage}[t]{0.47\columnwidth}
\centering
\includegraphics[width=\linewidth]{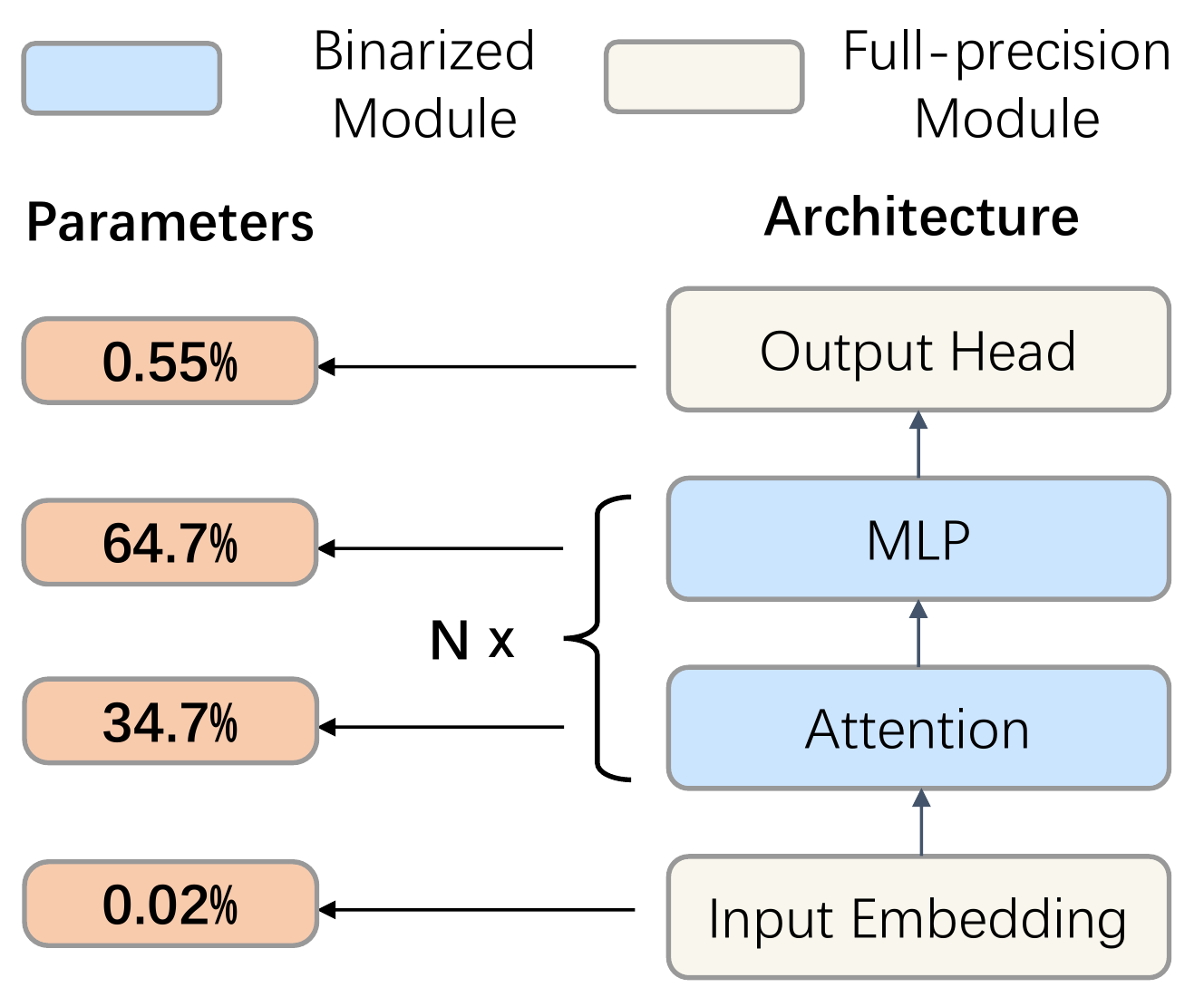}
\caption{\textbf{The architecture and parameters of Swin-T model.} MLP is difficult to binarize due to limited representational capacity and has far more parameters than other modules.}
\label{fig:cr}
\end{minipage}
\hfill
\begin{minipage}[t]{0.47\columnwidth}
\centering
\includegraphics[width=\linewidth]{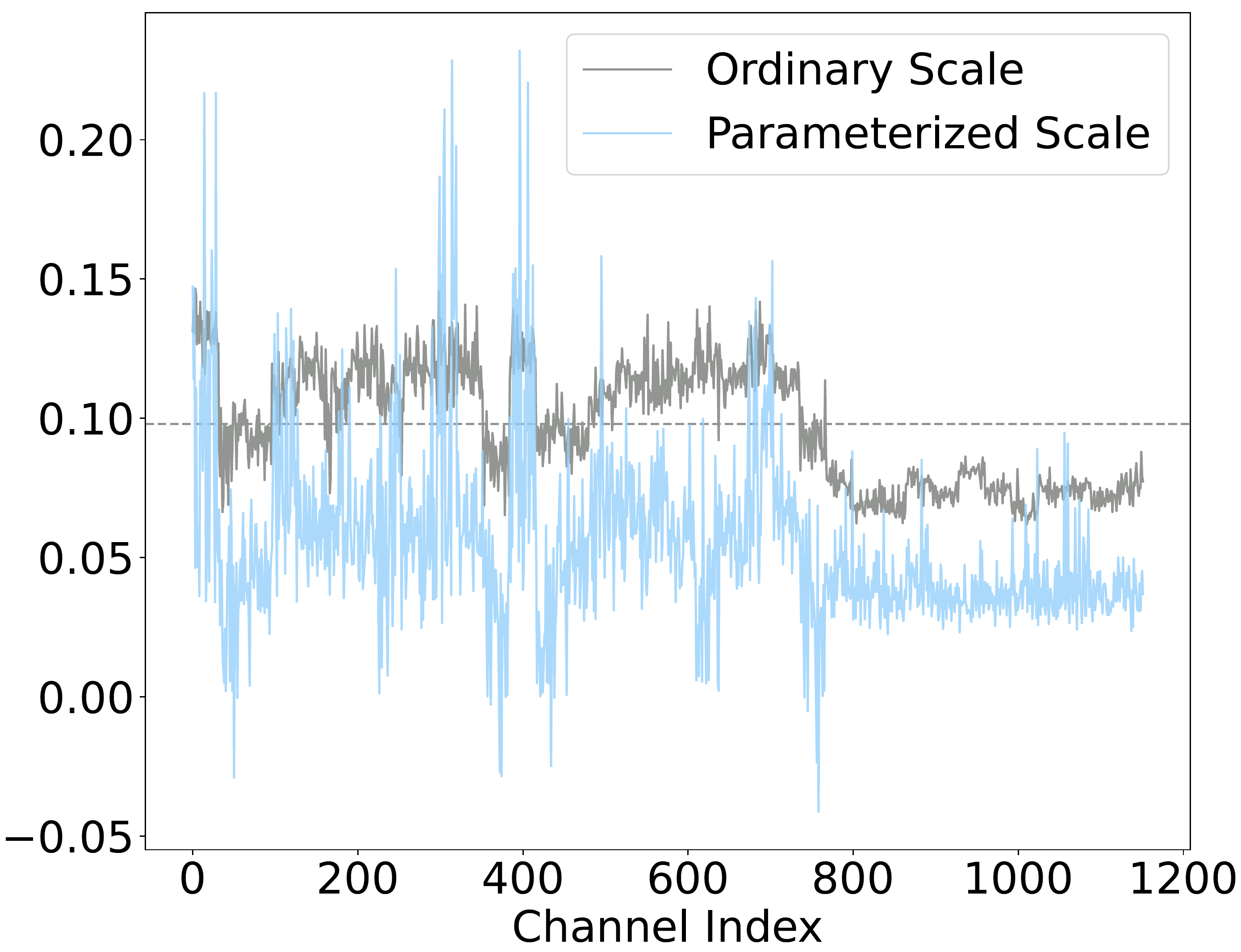}
\caption{\textbf{Ordinary and parameterized scaling factors in NesT-T model.} Data are collected from binarized Nest-T model. The dashed line represents the mean value of ordinary scaling factors.}
\label{fig:scale} 
\end{minipage}
\vspace{-0.5em}
\end{figure}

To explore the reasons, we present the architecture and parameters of Swin-T as an example in Figure~\ref{fig:cr}. As analyzed in Section~\ref{sec:intro}, the MLP modules are hard to quantize due to the limited representational capability of $1 \times 1$ convolutions
and have more parameters than the self-attention module. To tackle this problem, we propose Cross-layer Binarization (CLB), which is analogous to the previous two-step training scheme~\cite{zhuang2018towards,martinez2020training}, to decouple the quantization of self-attention module and MLP module to reduce mutual interference. Specifically, in the first stage, we keep MLPs to full precision and binarize all the self-attention modules with our SAB scheme. Then in the second stage, we binarize MLPs to get the final model. 


Compared with previous two-step training schemes that first binarize activations and then weights~\cite{martinez2020training,liu2020reactnet,bahri2021binary}, CLB is designed for Transformers to relieve the mutual interference and mitigate information loss mainly caused by binarizing MLPs. 

The experimental results demonstrate that using CLB brings more accuracy improvement than the traditional two-step training scheme, as we will show in Section~\ref{sec:ablacr}.

\subsubsection{Parameterized Weight Scales}
To further narrow the performance gap between the binarized model and the full-precision counterpart, an intuitive idea is to increase the representation ability of the binarized model. However, this usually results in increased computational complexity.

Motivated by SE-Net~\cite{hu2018squeeze}, channel-wise scaling factors can be regarded as the importance of each channel, rather than an approximation of its distribution. 
In order to preserve the model structure and complexity while enhancing its representation ability, we directly replace ordinary scaling factors (as defined in Eq.~(\ref{eq:scale})) with learnable parameters. The parameterized scaling factors could be optimized in conjunction with other network parameters via backward propagation during training. As shown in Figure~\ref{fig:scale}, the deviation of ordinary scaling factors across channels is small, indicating that the weight distribution of each channel is similar. In contrast, the parameterized scaling factors vary greatly from channel to channel, showing that it learns to pay more attention to specific channels and thus enhancing the representational capacity of the model. 
\section{Experiments on Image Classification}
\subsection{Implementation Details}
\noindent\textbf{Datasets and architectures.} We conduct image classification experiments on two standard benchmarks: TinyImageNet~\cite{wu2017tiny} and ImageNet (ILSVRC12)~\cite{deng2009imagenet}. 
The input resolution is $224 \times 224$. For data augmentation, we follow the settings in DeiT~\cite{pmlr-deit-touvron21a}, which are common practices in ViTs. 
To demonstrate the versatility of our method, we adopt three widely-used efficient architectures: DeiT~\cite{pmlr-deit-touvron21a}, Swin~\cite{liu2021swin} and NesT~\cite{zhang2022nested}.
All the blocks in Transformer models are binarized without exception. For binary attention modules, all weights and intermediate results including $\mathbf{Q}$, $\mathbf{K}$, $\mathbf{V}$ and projection layers, are binarized. For binary MLP modules, weights are binarized in all experiments. We leave the input embedding layer and output layer unbinarized as it is the common practice of BNNs~\cite{rastegari2016xnor}. 
The binary operations (BOPs) and floating-point operations (FLOPs) are counted separately and the reported total operations (OPs) are calculated by $\mathrm{OPs=BOPs/64+FLOPs}$, following~\cite{rastegari2016xnor, liu2020reactnet}.

\noindent\textbf{Training setup.} All experiments are implemented with PyTorch~\cite{paszke2019pytorch} and Timm~\cite{rw2019timm} library. The iteration number for coordinate descent is $N=5$. For both datasets, we employ Adam~\cite{kingma2014adam} optimizers without weight decay and train models for 300 epochs using a cosine annealing schedule with 5 epochs of warm-up. The initial learning rate is set to 5e-4. When training is split into two stages, we train 150 epochs at each stage to keep the number of total iterations the same. Knowledge distillation (KD)~\cite{hinton2015distilling} is used in all experiments. Specifically, we use the distribution loss proposed in ~\cite{liu2020reactnet} for optimization. Before training, all parameters are initialized with the pretrained model. 

\subsection{Comparison with SOTA methods}
\subsubsection{Evaluation on TinyImgnet} \label{sec:tinyimgnet}
\begin{table}[ht]
\caption{Comparisons of different network architectures on TinyImageNet. Here, ``FP'' means full-precision pretrained model, ``ATTN'' denotes attention module and ``W/A'' represents the number of bits used in weights or activations. 
}
\label{tinyimgnetresult}
\centering
\resizebox{0.85\columnwidth }{!}{
\begin{tabular}{ccccc}
\hline
Model                    & Method & \begin{tabular}[c]{@{}c@{}}ATTN\\ Bitwidth (W/A)\end{tabular} & \begin{tabular}[c]{@{}c@{}}MLP\\ Bitwidth (W/A)\end{tabular} & \begin{tabular}[c]{@{}c@{}}Top-1\\ Acc. (\%)\end{tabular}     \\ \hline
\multirow{8}{*}{Swin-T} & FP     & 32/32                                                         & 32/32                                                        & 80.57          \\
                        & BiBERT & 1/1                                                           & 1/1                                                          & 41.89          \\
                        & BiT    & 1/1                                                           & 1/1                                                          & 40.52          \\
                        & Ours   & 1/1                                                           & 1/1                                                          & \textbf{58.66} \\ \cline{2-5} 
                        & FP     & 32/32                                                         & 32/32                                                        & 80.57          \\
                        & BiBERT & 1/1                                                           & 1/32                                                         & 65.93          \\
                        & BiT    & 1/1                                                           & 1/32                                                         & 61.82          \\
                        & Ours   & 1/1                                                           & 1/32                                                         & \textbf{71.20} \\ \hline
\multirow{8}{*}{NesT-T} & FP     & 32/32                                                         & 32/32                                                        & 80.31          \\
                        & BiBERT & 1/1                                                           & 1/1                                                          & 32.39          \\
                        & BiT    & 1/1                                                           & 1/1                                                          & 34.72          \\
                        & Ours   & 1/1                                                           & 1/1                                                          & \textbf{52.21} \\ \cline{2-5} 
                        & FP     & 32/32                                                         & 32/32                                                        & 80.31          \\
                        & BiBERT & 1/1                                                           & 1/32                                                         & 49.53          \\
                        & BiT    & 1/1                                                           & 1/32                                                         & 46.43          \\
                        & Ours   & 1/1                                                           & 1/32                                                         & \textbf{69.83} \\ \hline
\end{tabular}
}
\end{table}

We begin our evaluation on the TinyImageNet, and the results are presented in Table~\ref{tinyimgnetresult}. Previous Transformer binarization methods~\cite{liu2022bit, qin2022bibert} exhibit significant accuracy degradation, which limits their practicality.  Remarkably, in some cases, the performance of BiT~\cite{liu2022bit} can be worse than BiBERT~\cite{qin2022bibert}, indicating that the learned thresholds $T$ may not be suitable for softmax attentions during inference or can even be detrimental. By contrast, our proposed SAB method dynamically determines the thresholds according to the distribution of softmax attention. For models with all weights and activations binarized, our approach can boost the Top-1 accuracy by almost $20\%$ ($52.21\%$ vs. $32.39\%$ for NesT-T), greatly narrowing the performance gap between binary and full-precision models. 
However, we still observed a large accuracy drop, which can be attributed to the significant contribution of MLP modules to the parameters and their limited representational capability. To enhance practicality, we keep the activations in MLP modules as full-precision.

The experimental results demonstrate that preserving activations as full-precision can significantly mitigate the performance degradation caused by fully-binarized MLPs ($69.83\%$ vs. $52.21\%$ for NesT-T), achieving a better trade-off between accuracy and complexity. In this case, the model size can still be compressed by 32$\times$, and the costly multiply-accumulate operations can be replaced by cheap accumulations. With this configuration, our method also outperforms previous methods BiBERT and BiT by large margins, with improvements of up to $20.3\%$ for Nest-T.

We also found that binary Swin-T outperforms binary NesT-T, while their full-precision models perform similarly. This difference may be attributed to the mask mechanism in the local window attention of Swin, which limits the number of elements greater than zero in the pre-softmax attention,
thereby aiding in model binarization. Nevertheless, our method provides substantial accuracy gains for most Transformers with standard self-attention (like NesT).
\subsubsection{Evaluation on ImageNet}
We further evaluate the effectiveness of our method on the ImageNet, and the results are shown in Table~\ref{imgnetresult}. To preserve the accuracy, we binarize all weights while leaving the activations in MLP modules full-precision. In all cases, our method obtains the best performance. For full-attention models (like DeiT), the accuracy is significantly lower than local-attention models, especially when the model capacity is insufficient (such as DeiT-T). We 
speculate
that the inductive bias of local information is very important for highly compressed BiViTs. For Swin-T, our method outperforms previous SOTA by a margin of $2.5\%$. For NesT-T, where the previous method even fails to converge, our method obtains a $68.7\%$ Top-1 accuracy. We also conduct experiments on larger ViTs like DeiT-B and Swin-S, achieving a highly competitive accuracy of up to $75.6\%$ over Swin-S. These results demonstrate the feasibility of BiViTs in 
visual tasks for the first time.

\begin{table}[t]
\small
\caption{Performance comparisons of different architectures on ImageNet.
``*'' denotes the model fails to converge. }
\label{imgnetresult}
\centering
\scalebox{0.9}{
\begin{tabular}{cccccc}
\hline
\begin{tabular}[c]{@{}c@{}}Base\\ Arch\end{tabular} & Model                      & Method & \begin{tabular}[c]{@{}c@{}}Size\\ (MB)\end{tabular} & \begin{tabular}[c]{@{}c@{}}OPs\\ (10\textasciicircum{}9)\end{tabular} & \begin{tabular}[c]{@{}c@{}}Top-1\\ Acc.(\%)\end{tabular} \\ \hline
\multirow{6}{*}{CNN}                                & \multirow{3}{*}{ResNet-18} & FP     & 46.8                                                & 1.83                                                                  & 69.6                                                     \\
                                                    &                            & AdaBin\footnotemark[1] & 5.56                                                & 0.18                                                                  & 63.1                                                     \\
                                                    &                            & IR-Net\footnotemark[2] & 5.56                                                & 0.99                                                                  & 66.5                                                     \\ \cline{2-6} 
                                                    & \multirow{3}{*}{ResNet-34} & FP     & 87.2                                                & 3.68                                                                  & 73.3                                                     \\
                                                    &                            & AdaBin\footnotemark[1] & 8.12                                                & 0.21                                                                  & 66.4                                                     \\
                                                    &                            & IR-Net\footnotemark[2] & 8.12                                                & 2.11                                                                  & 70.4                                                     \\ \hline
\multirow{24}{*}{Transformer}                       & \multirow{3}{*}{PVTv2-B1}  & FP     & 53.5                                                & 2.12                                                                  & 78.8                                                     \\
                                                    &                            & BiBERT & 11.3                                                & 0.95                                                                  & 64.7                                                     \\
                                                    &                            & Ours   & 11.3                                                & 0.95                                                                  & \textbf{67.3}                                            \\ \cline{2-6} 
                                                    & \multirow{3}{*}{CvT-13}    & FP     & 76.6                                                & 4.59                                                                  & 81.4                                                     \\
                                                    &                            & BiBERT & 7.11                                                & 1.73                                                                  & 55.4                                                     \\
                                                    &                            & Ours   & 7.11                                                & 1.73                                                                  & \textbf{64.9}                                            \\ \cline{2-6} 
                                                    & \multirow{3}{*}{DeiT-T}    & FP     & 22.8                                                & 1.26                                                                  & 71.7                                                     \\
                                                    &                            & BiBERT & 2.22                                                & 0.39                                                                  & 25.4                                                     \\
                                                    &                            & Ours   & 2.22                                                & 0.39                                                                  & \textbf{37.9}                                            \\ \cline{2-6} 
                                                    & \multirow{3}{*}{DeiT-B}    & FP     & 346.3                                               & 17.6                                                                  & 81.8                                                     \\
                                                    &                            & BiBERT & 16.8                                                & 5.81                                                                  & 67.5                                                     \\
                                                    &                            & Ours   & 16.8                                                & 5.81                                                                  & \textbf{69.6}                                            \\ \cline{2-6} 
                                                    & \multirow{3}{*}{Swin-T}    & FP     & 113.2                                               & 4.51                                                                  & 81.2                                                     \\
                                                    &                            & BiBERT & 12.6                                                & 1.62                                                                  & 68.3                                                     \\
                                                    &                            & Ours   & 12.6                                                & 1.62                                                                  & \textbf{70.8}                                            \\ \cline{2-6} 
                                                    & \multirow{3}{*}{Swin-S}    & FP     & 198.4                                               & 8.77                                                                  & 83.2                                                     \\
                                                    &                            & BiBERT & 15.4                                                & 3.04                                                                  & 74.0                                                     \\
                                                    &                            & Ours   & 15.4                                                & 3.04                                                                  & \textbf{75.6}                                            \\ \cline{2-6} 
                                                    & \multirow{3}{*}{Nest-T}    & FP     & 68.4                                                & 5.83                                                                  & 81.1                                                     \\
                                                    &                            & BiBERT & 8.96                                                & 2.49                                                                  & 0.27*                                                    \\
                                                    &                            & Ours   & 8.96                                                & 2.49                                                                  & \textbf{68.7}                                            \\ \cline{2-6} 
                                                    & \multirow{3}{*}{Nest-S}    & FP     & 153.4                                               & 10.4                                                                  & 83.3                                                     \\
                                                    &                            & BiBERT & 11.7                                                & 3.92                                                                  & 73.5                                                     \\
                                                    &                            & Ours   & 11.7                                                & 3.92                                                                  & \textbf{74.9}                                            \\ \hline
\end{tabular}}
\end{table}
\footnotetext[1]{CNNs with binary weights and activations.}
\footnotetext[2]{CNNs with binary weights and full-precision activations.}

\subsection{Comparison with Binary CNNs}
We present a comparison of the model size, total operations, and accuracy between binary ResNet ~\cite{he2016deep} and BiViTs, as shown in Table~\ref{imgnetresult}. CNNs with their convolutional inductive bias and fewer $1 \times 1$ convolution layers can achieve good performance with minimal parameters and operations, surpassing similarly sized BiViTs such as DeiT-T. However, as model capacity increases, the advantages of Transformers become more apparent. The global receptive field and attention mechanism in Transformers provide strong representational capability, resulting in higher accuracy for full-precision ViT models than ResNet models with similar parameters. As the information of the pretrained model is inherited, a stronger teacher model can significantly improve BiViT's training. For instance, binary Swin-S has similar OPs as full-precision ResNet-34, yet it achieves better accuracy ($75.6\%$ vs. $73.3\%$) and reduces model size by $5.66\times$. 
Moreover, we measure the latency of matrix multiplication operations in BiViTs and full-precision models using an RTX3090 GPU, as shown in Table~\ref{tab:latency}.
For floating-point (FP) operations, we utilize cuDNN~\cite{chetlur2014cudnn}, while binary operations are implemented using BTC-BNN~\cite{tcbnn}. 
In our default setting where attention modules are binarized while the MLP modules retain FP activations, we observe a $1.99\times$ reduction in latency compared to its full-precision counterpart on Swin-T. Moreover, when we fully binarize the MLP modules as well, the speedup further increases to $4.39\times$. However, due to the long-standing optimization of convolution implementations, such as Winograd~\cite{winograd}, the current latency of BiViTs is slightly higher than that of FP CNNs with similar OPs, such as binary Swin-S model and FP ResNet-34.

\begin{table}[h] \small
\caption{Latency comparisons with different models on ImageNet.}
\label{tab:latency}
\centering
\resizebox{0.9\columnwidth }{!}{
\begin{tabular}{cccccc}
\hline
Model                   & Method & \begin{tabular}[c]{@{}c@{}}Size\\ (MB)\end{tabular} & \begin{tabular}[c]{@{}c@{}}OPs\\ (10\textasciicircum{}9)\end{tabular} & \begin{tabular}[c]{@{}c@{}}Latency\\ (ms)\end{tabular} & \begin{tabular}[c]{@{}c@{}}Top-1\\ Acc. (\%)\end{tabular} \\ \hline
ResNet-18               & FP     & 46.8                                                & 1.83                                                                  & 0.91                                                   & 69.6                                                      \\ \hline
ResNet-34               & FP     & 87.2                                                & 3.68                                                                  & 2.46                                                   & 73.3                                                      \\ \hline
\multirow{3}{*}{Swin-T} & FP     & 113.2                                               & 4.51                                                                  & 3.03                                                  & 81.2                                                      \\
                        & Ours\footnotemark[3]   & 12.6                                                & 1.62                                                                  & 1.52                                                   & 70.8                                                      \\
                        & Ours\footnotemark[4]   & 12.6                                                & 0.27                                                                  & 0.69                                                   & -                                                         \\ \hline
\multirow{3}{*}{Swin-S} & FP     & 198.4                                               & 8.77                                                                  & 5.42                                                  & 83.2                                                      \\
                        & Ours\footnotemark[3]   & 15.4                                                & 3.04                                                                  & 2.91                                                  & 75.6                                                      \\
                        & Ours\footnotemark[4]   & 15.4                                                & 0.34                                                                  & 1.16                                                   & -                                                         \\ \hline
\multirow{3}{*}{Nest-T} & FP     & 68.4                                                & 5.83                                                                  & 3.50                                                   & 81.1                                                      \\
                        & Ours\footnotemark[3]   & 8.96                                                & 2.49                                                                  & 2.08                                                   & 68.7                                                      \\
                        & Ours\footnotemark[4]   & 8.96                                                & 1.12                                                                  & 1.21                                                   & -                                                         \\ \hline
\multirow{3}{*}{Nest-S} & FP     & 153.4                                               & 10.4                                                                  & 5.88                                                  & 83.3                                                      \\
                        & Ours\footnotemark[3]   & 11.7                                                & 3.92                                                                  & 3.46                                                  & 74.0                                                      \\
                        & Ours\footnotemark[4]   & 11.7                                                & 1.19                                                                  & 1.70                                                   & -                                                         \\ \hline
\end{tabular} }
\end{table}
\footnotetext[3]{Transformers with binary weights, binary activations within attentions and full-precision activations within MLPs.}
\footnotetext[4]{Transformers with binary weights and activations.}

\subsection{Ablation Studies}
\subsubsection{Effect of Softmax-aware Binarization}  \label{abla:sab}
First, we conducted ablation experiments over Swin-T and NesT-T models on TinyImageNet to prove the effectiveness of the proposed SAB scheme. To eliminate the impact of MLPs, we keep them full-precision and only binarize attention modules. As shown in Table~\ref{abla:attn}, our SAB consistently narrows the accuracy gap between the full-precision teacher and the binary student network, indicating successful suppression of the quantization error in the self-attention module. Incorporating $\mathrm{Softmax}$ for gradient approximation brings an accuracy improvement of 2.02\%, because it mitigates the mismatch issue between backpropagation using STE and the $\mathrm{Softmax}$ operation in the forward pass.

\begin{table}[t]
\caption{Ablation study on Softmax-aware Binarization. $\dagger$ denotes the network not considering $\mathrm{Softmax}$ in the backward pass.}
\label{abla:attn}
\tiny
\centering
\resizebox{0.8\columnwidth}{!}{
\begin{tabular}{cccc}
\hline
Model                    & Method                       & \begin{tabular}[c]{@{}c@{}}ATTN\\ Bitwidth (W/A)\end{tabular} & \begin{tabular}[c]{@{}c@{}}Top-1\\ Acc. (\%)\end{tabular}     \\ \hline
\multirow{3}{*}{Swin-T} & FP                           & 32/32                                                         & 80.57          \\
                        & BiBERT                       & 1/1                                                           & 73.39          \\
                        & + SAB & 1/1                                                           & \textbf{74.62} \\ \hline
\multirow{4}{*}{NesT-T} & FP                           & 32/32                                                         & 80.31          \\
                        & BiBERT                       & 1/1                                                           & 68.51          \\
                         & + SAB$\dagger$                    & 1/1                                                           & 68.71          \\
                        & + SAB & 1/1                                                           & \textbf{70.73} \\ \hline
\end{tabular}}
\end{table}

Also, we conduct experiments to verify the impact of $\beta$ estimation (see Eq.~(\ref{eq:alpha})) on the accuracy of the model. The results in Table~\ref{tab:thresh} show that the model is not sensitive to $\beta$ within a reasonable interval (about $0.25$ to $0.45$). By default, $\beta$ is set to $0.25$ to enable efficient bit-shift operation in Eq.~(\ref{eq:alpha}). 
However, the accuracy of the model decreases when $\beta$ is too small (less than $0.2$). This indicates that as the thresholds become too small, our SAB is less effective and too many attention scores are activated.

\begin{table}[]
\small
\caption{Comparisons of binary attention's performance under different thresholds. The experiment is conducted over NesT-T model on TinyImageNet.}
\label{tab:thresh}
\centering
\begin{tabular}{cc}
\hline
Method            & Top-1 Acc. (\%) \\ \hline
FP                & 80.31          \\
BiBERT            & 68.51          \\
Ours ($\beta$=0.20) & 69.18          \\
Ours ($\beta$=0.25) & 70.73          \\
Ours ($\beta$=0.35) & 70.81          \\
Ours ($\beta$=0.45) & 70.68          \\ \hline
\end{tabular}
\end{table}

\subsubsection{Effect of Information Preservation} \label{sec:ablacr}
\begin{figure}
\centering
\includegraphics[width=0.9\linewidth]{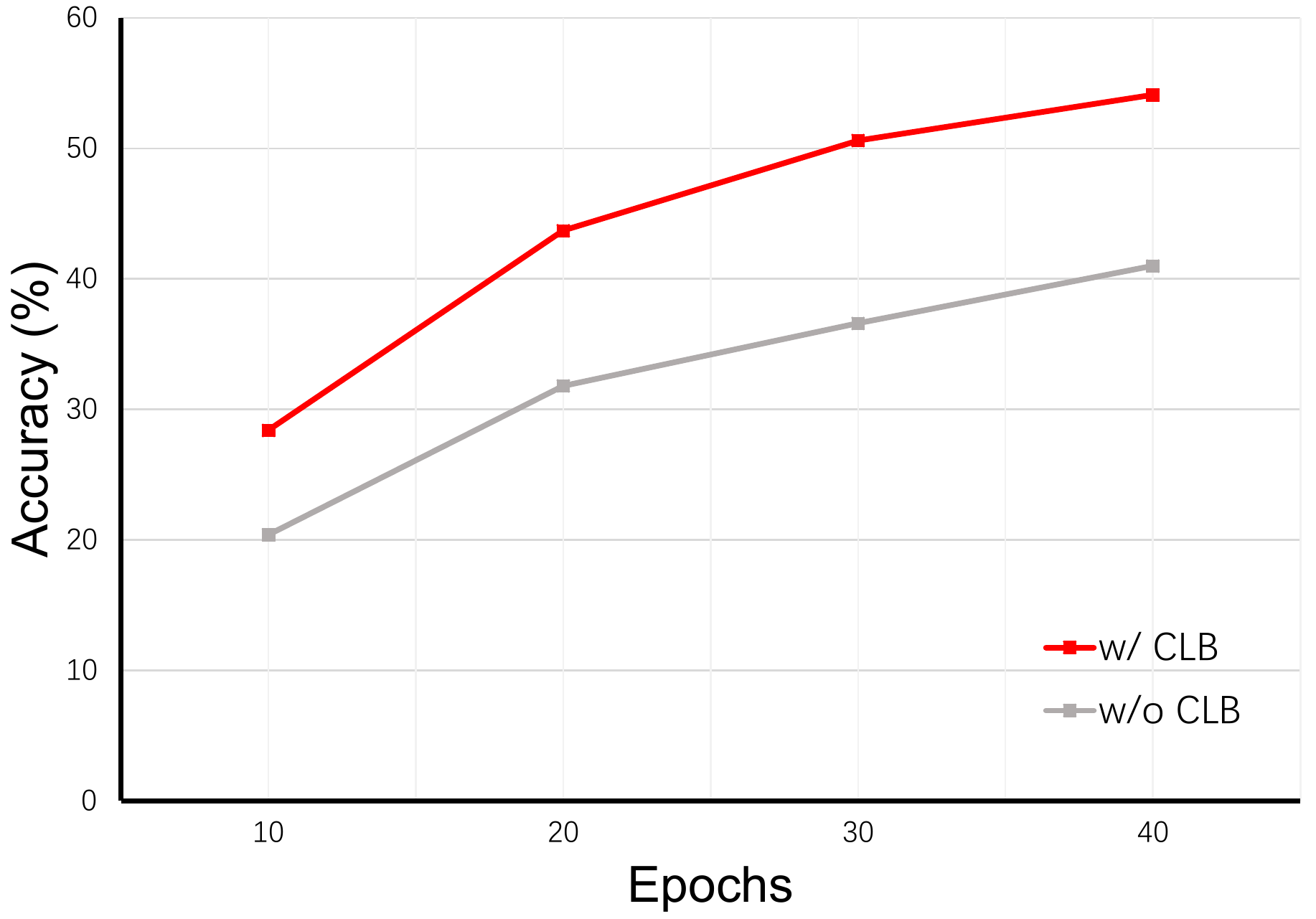}
\caption{\textbf{Training accuracy curves with less training epochs.} 
    The experiment is conducted over NesT-T model on TineImageNet.}
\label{fig:cr_training} 
\end{figure}

To demonstrate the effectiveness of our proposed CLB training scheme, we conducted experiments comparing the accuracy of binary Nest-T trained with CLB to those trained using traditional two-step schemes~\cite{martinez2020training}, as well as those trained directly with one-step training scheme on TinyImageNet. The activations in MLP modules remain full-precision. The results, shown in Figure~\ref{fig:cr_training} and Table~\ref{abla:cr}, demonstrate that using CLB can accelerate the convergence process of the BiViTs and significantly improve the accuracy by $11.6\%$ compared to one-step training. Furthermore, CLB outperforms the conventional two-step training scheme by $4.1\%$, highlighting its strong ability to retain pretrained information. 
We expect that 
our CLB and the conventional two-step training
methods 
can be combined to further improve the accuracy since they are orthogonal, but will leave it for future work. 

\begin{table}[h]
\caption{Ablation study on Cross-layer Binarization. Here, ``TS'' denotes traditional two-step training scheme. }
\label{abla:cr}
\centering
\resizebox{0.9\columnwidth }{!}{
\begin{tabular}{cccc}
\hline
Method        & \begin{tabular}[c]{@{}c@{}}ATTN\\ Bitwidth (W/A)\end{tabular} & \begin{tabular}[c]{@{}c@{}}MLP\\ Bitwidth (W/A)\end{tabular} & \begin{tabular}[c]{@{}c@{}}Top-1\\ Acc. (\%)\end{tabular}    \\ \hline
FP            & 32/32                                                         & 32/32                                                        & 80.31          \\
Ours (w/o CLB) & 1/1                                                           & 1/32                                                         & 58.20          \\
Ours (w/ TS)  & 1/1                                                           & 1/32                                                         & \textbf{65.64} \\
Ours (w/ CLB)  & 1/1                                                           & 1/32                                                         & \textbf{69.83} \\ \hline
\end{tabular} }
\vspace{-0.5em}
\end{table}



For PWS, it can be applied to both self-attention and MLP modules. To show the improvement brought by PWS, we conduct experiments starting with
a BiBERT-based baseline and then changing scaling factors to learnable parameters in self-attention and MLP modules separately. As shown in Table~\ref{abla:adap}, PWS is effective in both modules, resulting in an accuracy improvement of 4.2\% and 1.3\%, respectively. Therefore, we use PWS in both modules by default if they are binarized.

\begin{table}[h]\footnotesize
\caption{Ablation study on Parameterized Weight Scales. The experiment is conducted over Nest-T model on TinyImageNet.}
\label{abla:adap}
\centering
\resizebox{0.9\columnwidth }{!}{
\begin{tabular}{cccc}
\hline
Method       & \begin{tabular}[c]{@{}c@{}}ATTN\\ Bitwidth (W/A)\end{tabular} & \begin{tabular}[c]{@{}c@{}}MLP\\ Bitwidth (W/A)\end{tabular} & Top-1 (\%)     \\ \hline
FP           & 32/32                                                         & 32/32                                                        & 80.31          \\
BiBERT       & 1/1                                                           & 32/32                                                        & 68.51          \\
+PWS & 1/1                                                           & 32/32                                                        & \textbf{72.75}     \\ \hline
FP           & 32/32                                                         & 32/32                                                        & 80.31          \\
BiBERT       & 32/32                                                         & 1/1                                                          & 70.02          \\
+PWS & 32/32                                                         & 1/1                                                          & \textbf{71.35} \\ \hline
\end{tabular} }
\end{table}

\section{Experiments on Object Detection}
In this section, we present the results of object detection and instance segmentation experiments over Swin-T on COCO 2017~\cite{lin2014microsoftcoco} validation set. Currently, we only binarize the self-attention modules in the backbone. The experiments are implemented with classic object detection frameworks Mask R-CNN~\cite{he2017mask} and Cascade Mask R-CNN~\cite{cai2018cascade}. For training strategy and hyper-parameters, we follow the implementation in Swin~\cite{liu2021swin}, which takes ImageNet pretrained model as initialization and only trains for $12$ epochs.

Table~\ref{tab:coco} compares the results of different binarization methods on both frameworks. When evaluated with the Mask R-CNN framework, the proposed method improves the performance over BiBERT by $1.4\%$ and $1.2\%$ mAP on the object detection and instance segmentation tasks, respectively.
For the Cascade Mask R-CNN framework, the performance improvement brought by our method is even greater, achieving a competitive $40.8\%$ mAP on object detection. More results on COCO are available in the supplementary material.

\begin{table}[ht]
\small
\caption{Comparisons of different methods and backbones on COCO 2017 validation set. Here, ``FP'' represents the full-precision pretrained model.
}
\label{tab:coco}
\centering
\resizebox{1.0\columnwidth }{!}{
\begin{tabular}{cccccc}
\hline
Framework                           & Method & Task                                                                             & AP            & AP$_{50}$     & AP$_{75}$     \\ \hline
\multirow{6}{*}{Mask R-CNN}         & FP     & \multirow{3}{*}{\begin{tabular}[c]{@{}c@{}}Object\\ Detection\end{tabular}}      & 43.7          & 66.6          & 47.7          \\
                                    & BiBERT &                                                                                  & 32.0          & 53.9          & 33.7          \\
                                    & Ours   &                                                                                  & \textbf{33.4} & \textbf{55.0} & \textbf{35.2} \\ \cline{2-6} 
                                    & FP     & \multirow{3}{*}{\begin{tabular}[c]{@{}c@{}}Instance\\ Segmentation\end{tabular}} & 39.8          & 63.3          & 42.7          \\
                                    & BiBERT &                                                                                  & 30.4          & 51.0          & 31.9          \\
                                    & Ours   &                                                                                  & \textbf{31.6} & \textbf{51.7} & \textbf{33.4} \\ \hline
\multirow{6}{*}{Cascade Mask R-CNN} & FP     & \multirow{3}{*}{\begin{tabular}[c]{@{}c@{}}Object\\ Detection\end{tabular}}      & 48.1          & 67.1             & 52.2             \\
                                    & BiBERT &                                                                                  & 39.2             & 57.3             & 42.5             \\
                                    & Ours   &                                                                                  & \textbf{40.8}           & \textbf{59.2}             & \textbf{44.1}             \\ \cline{2-6} 
                                    & FP     & \multirow{3}{*}{\begin{tabular}[c]{@{}c@{}}Instance\\ Segmentation\end{tabular}} & 41.7          & 64.4             & 45.0             \\
                                    & BiBERT &                                                                                  & 34.5             & 54.5             & 36.8             \\
                                    & Ours   &                                                                                  & \textbf{35.7}            & \textbf{56.5}             & \textbf{38.2}             \\ \hline
\end{tabular}
}
\end{table}
\vspace{-1em}
\section{Conclusion}
In this paper, we have proposed 
to tackle two fundamental challenges with customized solutions
for BiViTs, and have successfully applied BiViTs to visual tasks for the first time. 
To deal with
the long-tailed distribution of softmax attention, we have proposed the Softmax-aware Binarization for self-attention, the core module of Transformers, which greatly reduces the quantization error. To preserve information from pretrained model, we have proposed the Cross-layer Binarization scheme that decouples the quantization of self-attention and MLPs. Moreover, we have introduced Parameterized Weight Scales to enhance representational ability of BiViTs. 
Our BiViT has achieved significant accuracy improvement over SOTA on the image classification task, with up to 75.6\% Top-1 accuracy on ImageNet over Swin-S model. 
We have also conducted extensive experiments on COCO object detection and instance segmentation, demonstrating that BiViTs can be extended to downstream tasks. In the future, we will further investigate methods to narrow the gap between BiViTs and their full-precision counterparts.

~\\
\noindent \textbf{Acknowledgement}
This work was supported by the National Key Research and Development Program of China (2022YFC3602601) and Key Research and Development Program of Zhejiang Province of China (2021C02037).

{\small
\bibliographystyle{ieee_fullname}
\bibliography{reference}
}

\newpage
\onecolumn

\begin{center}
	{
		\Large{\textbf{Appendix}}
	}
\end{center}
\appendix
\setcounter{section}{0}
\setcounter{equation}{0}
\setcounter{figure}{0}
\setcounter{table}{0}

\renewcommand\thesection{\Alph{section}}
\renewcommand\thefigure{\Alph{figure}}
\renewcommand\thetable{\Alph{table}}
\renewcommand{\theequation}{\Alph{equation}}
\section{More experimental results on COCO}
In this section, we present additional experimental results of BiViT on the COCO 2017 validation set. As shown in Table~\ref{tab:appendix_coco}, our method outperforms the competitive BiBERT by a large margin in all metrics.
\begin{table}[h]
\caption{Comparisons of different methods and backbones on COCO 2017 validation set.}
\label{tab:appendix_coco}
\centering
\begin{tabular}{ccccccccc}
\hline
Framework                           & Method & Task                                                                             & AP            & AP$_{50}$     & AP$_{75}$     & AP$_{S}$      & AP$_{M}$      & AP$_{L}$      \\ \hline
\multirow{6}{*}{Mask R-CNN}         & FP     & \multirow{3}{*}{\begin{tabular}[c]{@{}c@{}}Object\\ Detection\end{tabular}}      & 43.7          & 66.6          & 47.7          & 28.5          & 47.0          & 57.3          \\
                                    & BiBERT &                                                                                  & 32.0          & 53.9          & 33.7          & 19.6          & 34.1          & 41.3          \\
                                    & Ours   &                                                                                  & \textbf{33.4} & \textbf{55.0} & \textbf{35.2} & \textbf{20.8} & \textbf{35.6} & \textbf{42.0} \\ \cline{2-9} 
                                    & FP     & \multirow{3}{*}{\begin{tabular}[c]{@{}c@{}}Instance\\ Segmentation\end{tabular}} & 39.8          & 63.3          & 42.7          & 24.2          & 43.1          & 54.6          \\
                                    & BiBERT &                                                                                  & 30.4          & 51.0          & 31.9          & 16.7          & 32.5          & 41.4          \\
                                    & Ours   &                                                                                  & \textbf{31.6} & \textbf{51.7} & \textbf{33.4} & \textbf{18.3} & \textbf{34.0} & \textbf{42.1} \\ \hline
\multirow{6}{*}{Cascade Mask R-CNN} & FP     & \multirow{3}{*}{\begin{tabular}[c]{@{}c@{}}Object\\ Detection\end{tabular}}      & 48.1          & 67.1          & 52.2          & 30.4          & 51.5          & 63.1          \\
                                    & BiBERT &                                                                                  & 39.2          & 57.3          & 42.5          & 23.7          & 41.3          & 51.1          \\
                                    & Ours   &                                                                                  & \textbf{40.8} & \textbf{59.2} & \textbf{44.1} & \textbf{25.2} & \textbf{43.3} & \textbf{52.8} \\ \cline{2-9} 
                                    & FP     & \multirow{3}{*}{\begin{tabular}[c]{@{}c@{}}Instance\\ Segmentation\end{tabular}} & 41.7          & 64.4          & 45.0          & 41.7          & 64.4          & 45.0          \\
                                    & BiBERT &                                                                                  & 34.5          & 54.5          & 36.8          & 19.1          & 36.5          & 46.6          \\
                                    & Ours   &                                                                                  & \textbf{35.7} & \textbf{56.5} & \textbf{38.2} & \textbf{20.1} & \textbf{38.2} & \textbf{48.5} \\ \hline
\end{tabular}
\end{table}



\end{document}